\documentclass[12pt]{article}

\usepackage{amsmath,amsthm, amsfonts, amssymb, amsxtra,amsopn}
\usepackage{pgfplots}
\pgfplotsset{compat=1.13}
\usepackage{pgfplotstable}
\usepackage{graphicx}
\usepackage{multirow}
\usepackage{tabu}
\usepackage{longtable}
\usepackage{booktabs}
\usepackage{listings}
\usepackage{cmap}
\usepackage{colortbl}
\usepackage{epsfig}
\usepackage[algo2e]{algorithm2e}
\usepackage{algorithm} 
\usepackage[noend]{algpseudocode} 

\usepackage{mdframed}
\usepackage{array} 
\usepackage[export]{adjustbox} 

\usepackage{tabularx}
\newcolumntype{Y}{>{\centering\arraybackslash}X}

\usepackage{subcaption}

\PassOptionsToPackage{hyphens}{url}

\usepackage{hyperref}
\hypersetup{colorlinks=true,linkcolor=black,citecolor=black,urlcolor=blue,filecolor=black}
\hypersetup{pdfpagemode=UseNone,pdfstartview=}

\usepackage[tableposition=top]{caption}
\usepackage{enumitem}
\setlist[itemize]{noitemsep, topsep=0pt}

\usepackage{apacite}

\advance\oddsidemargin by -0.25in
\advance\textwidth by 0.5in

\advance\topmargin by -0.3in
\advance\textheight by 0.6in

\long\def\symbolfootnotetext[#1]#2{\begingroup%
\def\thefootnote{\fnsymbol{footnote}}\footnotetext[#1]{#2}\endgroup}

\pgfkeys{
    /pgf/number format/fixed zerofill=true }
    
\pgfplotstableset{
    color cells/.code={%
        \pgfqkeys{/color cells}{#1}%
        \pgfkeysalso{%
            postproc cell content/.code={%
                \begingroup
                \pgfkeysgetvalue{/pgfplots/table/@preprocessed cell content}\value
\ifx\value\empty
\endgroup
\else
                \pgfmathfloatparsenumber{\value}%
                \pgfmathfloattofixed{\pgfmathresult}%
                \let\value=\pgfmathresult
                \pgfplotscolormapaccess
                    [\pgfkeysvalueof{/color cells/min}:\pgfkeysvalueof{/color cells/max}]%
                    {\value}%
                    {\pgfkeysvalueof{/pgfplots/colormap name}}%
                \pgfkeysgetvalue{/pgfplots/table/@cell content}\typesetvalue
                \pgfkeysgetvalue{/color cells/textcolor}\textcolorvalue
                \toks0=\expandafter{\typesetvalue}%
                \xdef\temp{%
                    \noexpand\pgfkeysalso{%
                        @cell content={%
                            \noexpand\cellcolor[rgb]{\pgfmathresult}%
                            \noexpand\definecolor{mapped color}{rgb}{\pgfmathresult}%
                            \ifx\textcolorvalue\empty
                            \else
                                \noexpand\color{\textcolorvalue}%
                            \fi
                            \the\toks0 %
                        }%
                    }%
                }%
                \endgroup
                \temp
\fi
            }%
        }%
    }
}

\title{Convolutional Neural Networks for Image Spam Detection}

\author{Tazmina Sharmin\footnotemark[1]\ \ \ 
Fabio Di Troia\footnotemark[1]\ \ \
Katerina Potika\footnotemark[1]\ \ \
Mark Stamp\footnotemark[1]\,\,\footnotemark[2]}

\begin{document}

\symbolfootnotetext[1]{Department of Computer Science, San Jose State University}
\symbolfootnotetext[2]{mark.stamp$@$sjsu.edu}

\maketitle

\abstract
Spam can be defined as unsolicited bulk email. In an
effort to evade text-based filters, spammers sometimes
embed spam text in an image, which is referred to as
image spam. In this research, we consider
the problem of image spam detection, based on image analysis.
We apply convolutional neural networks (CNN) to this problem,
we compare the results obtained using CNNs 
to other machine learning techniques,
and we compare our results to previous related work. We consider both 
real-world image spam and challenging
image spam-like datasets. Our results improve on previous work
by employing CNNs based on a novel 
feature set consisting of a combination of the raw image 
and Canny edges.

\section{Introduction}

Electronic mail or email is the most popular communication medium in the world~\cite{email_use}. 
As of~2015, the number of email users was~2.6 billion, while in~2019 this number will rise to 
approximately~2.9 billion, with more than one-third of world population using email to 
exchanging messages~\cite{email_stat}.

However, the effectiveness of email service is often reduced by spam. Spam is unwanted email with 
a commercial, fraudulent, or malicious purpose. As email usage has increased, the number of spam 
messages has also increased. Robust text-based filters have been developed to deal with spam. 
In an effort to evade such filters, spammers sometimes use image spam, that is, spammers 
insert their messages into images~\cite{image_spam_hunter}.

Previous research into image spam detection has shown that some types of image spam can be 
detected with high accuracy. For example, in the work presented 
in~\cite{image_spam,spam_svm_chavda}, a wide variety 
of image properties are extracted and images are classified as spam or ham (i.e., non-spam images) 
based on machine learning techniques. However, some challenging types of image spam are difficult 
to detect using such techniques. 

In this research, we conduct experiments to determine the effectiveness of various machine 
learning algorithms for image spam detection.
We consider a wide variety machine learning algorithms---including neural network based
techniques---over several image spam and image spam-like datasets. 
Overall, we find that a novel application of convolutional neural networks (CNN) performs
best and, in fact, this approach outperforms techniques considered in previous work.

The remainder of this paper is organized as follows. In Section~\ref{sect:background} we discuss 
relevant background topics and relevant previous work. Section~\ref{sect:mltechniques} provides an 
overview of the machine learning and neural network algorithms considered in this research. 
Section~\ref{sect:experiments} presents details on our various implementations and experimental results. 
Finally, Section~\ref{sect:conclusion} concludes the paper, and we briefly discuss possible avenues 
for future work.

\section{Background}\label{sect:background}

Spamming consists of sending unsolicited messages to a large number of users in an arbitrary manner. 
Spam has a wide variety of uses, ranging from advertising to online deception and other fraudulent activities. 
Since sending spam messages via email has little or no cost, email spam can be economically viable. 
In this section, we discuss spam in general, image spam in particular,
and we consider related work.

\subsection{Types of Spam}

In addition to email spam, there are other types of spam applicable to different means of communication. 
For example, mobile phone messaging spam is common, as well as web search engine spam 
and social networking spam~\cite{spam_filtering}.

Email spam is the most prevalent form of spam. In email spam, messages are sent to
a large number of email addresses. Such spam messages can include product advertisements, 
links to phishing websites, or links to malware installers. 
Historically, email spam contained only text messages. 
As text-based filters improved, image-based spam email emerged as a way to
bypass such filters~\cite{image_spam}. 
There are numerous other forms of email spam, including so-called
blank spam, which has no message in the email and is used to collect legitimate email addresses. 

Mobile phone message (SMS) spam refers to junk message sent to mobile phones. 
Such messages are inconvenient to mobile phone users, 
but since there are costs associated with SMS spam, 
it is less common than email spam~\cite{image_spam}.

Search engine spam refers to measures that attempt to affect the position of a website after a query. 
As a countermeasure, when a website is detected as having search engine spam, 
the site is marked and penalized.  One survey found 
that~51.3\% website hacks were related to search engine spam~\cite{web_hack}. 

Social spam aims at social networking websites such as Facebook and Twitter. 
One technique for social spamming consists of creating a fake account in a social application,
which is then used to hack into valid user accounts. These fake accounts are used to send 
bulk messages or malicious links, with the intent to harm. As social networking sites have become 
more popular, social spamming activities such as clickbaiting or likejacking have become 
more common~\cite{social_spam}.  

Gaming spam consists of sending messages in bulk to players using a common chat room or public discussion area. 
Spammers might target users who like gaming so as to sell gaming items for real-world money or in-game currency.

\subsection{Image Spam}

\begin{figure}[!htb]
\begin{center}
  \includegraphics[scale=0.55]{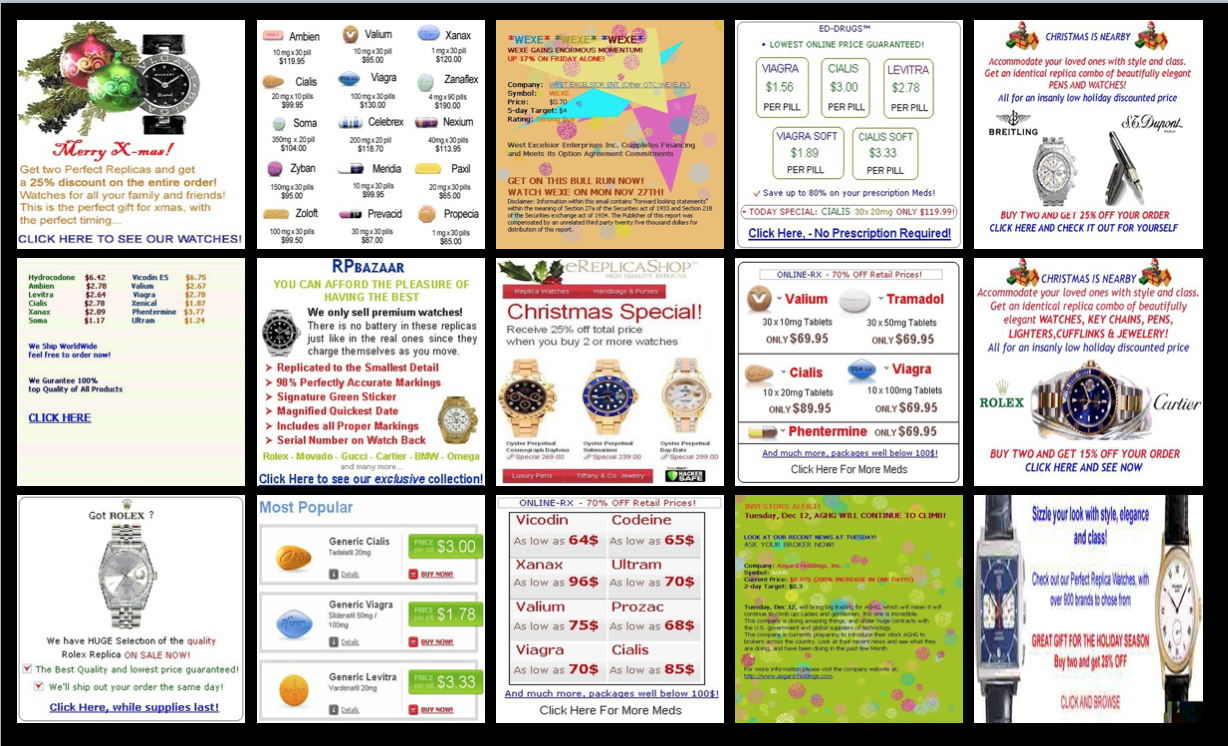}
  \caption{Examples of image spam~\protect\cite{image_spam_hunter}}\label{fig:spamExamples}
\end{center}
\end{figure}

Image spam is a subclass of email spam. As mentioned above, image spam emerged
as an obfuscation technique to evade text-based spam filters. Image spam 
is typically used to advertise products, deceive users to gain personal data, or to deliver malicious 
software~\cite{spam_filtering}. It is more challenging to detect image spam as compared to
text-based spam and image-based obfuscation techniques can be used to create image spam
that is even more challenging than that typically seen in practice~\cite{image_spam,spam_svm_chavda}.  
Examples of real-world image spam are given in Figure~\ref{fig:spamExamples}.

Image spam has evolved over time and can take several forms to bypass the conventional anti-spam 
techniques. The images used for spam can include text-only images, sliced images,
and randomized images, as discussed below.

The first generation image spam consisted of text-only images. 
Such images contain pure text embedded into an essentially blank image. 
These images look like text email, but are actually images. 
Optical character recognition (OCR) can be employed to extract such
text, at which point traditional text-based filters can be applied.

A sliced image consists of multiple images merged together in a jigsaw puzzle manner. 
This type of spam image is challenging to detect, and the combined image often 
passes through image spam filters.

Randomized images refers to randomization of the image pixels. To make a randomized image, 
spammers make changes to the individual pixel in the image. As a result, it may be difficult
to distinguish the randomized image from the original image. The changes made usually do 
not substantially affect the appearance of the image, but will alter hash values, and can even
influence the results of OCR based detection techniques.

\subsubsection{Image Spam Filtering}

Here, we discuss three general approaches to spam detection. Specifically, we consider 
header based, content based, and general non-content based techniques.
Note that these techniques are not mutually exclusive.

\begin{description}
\item[Header based]--- An email header consists of data about the sender and the receiver, 
including the sender's email address, date, from, to, and so on. The header fields of an email 
contain valuable information that may be useful in distinguishing spam and 
non-spam---email header attributes have been successfully used to train  
models to detect spam~\cite{header_filter}. Such techniques are 
applicable to image spam.

\item[Content based]--- Content based filters may, for example, check an email for 
particular keywords that are usually found in the body of spam messages. Typically, 
the body of an email carries the actual information to be delivered.
For image spam, OCR can be used to extract words that are then passed 
to a content based filter~\cite{apache}. 

\item[Non-content based]--- Non-content based techniques for image spam
rely on a direct analysis of image features, such as color properties, edge features, and so on. 
The goal is to use such image features to distinguish ham images from spam images. 

\end{description}

\subsection{Related Work}

Gao et al.~\cite{image_spam_hunter} propose an image spam detection scheme
that relies on a probabilistic boosting tree algorithm. 
Feature engineering based on the color and histogram of orientated gradient (HOG) 
are used to generate feature vectors for this learning algorithm. 
These authors obtain an accuracy of~0.8944.

Kumaresan et al.~\cite{spam_knn} propose a technique for detecting image spam based on 
color features, and using the $k$-nearest neighbor ($k$-NN) algorithm. Specifically, the
authors rely on the RGB and HSV histograms as features. 
In this research, a straightforward $k$-NN classifier yields
an accuracy of~0.945.

In the research by Annadatha et al.~\cite{image_spam}, support vector machines (SVM) have 
been applied to a set of~21 image features. Using feature selection based on linear SVM weights,
the authors are able to achieve an accuracy rate of~0.97 with a relatively small set of features.
The authors provide a challenge dataset
that could serve as image spam, but is much more difficult to detect, 
as compared to the real-world image spam.

Chavda et al.~\cite{spam_svm_chavda} conduct two sets of experiments with SVM and image processing. 
The authors use an extensive set of~41 image features and they achieve~0.97 and~0.98 accuracy 
on two publicly available datasets. These authors also provide a challenge dataset, which is shown to
be even more difficult to detect than that developed in~\cite{image_spam}.

Aiwan et al.~\cite{spam_cnn_aiwan} propose an image spam filtering method based on 
convolutional neural network (CNN).  
The proposed system uses data augmentation and achieves an accuracy improvement,
as compared to selected examples of previous work.
Kumar et al.~\cite{KumarDeep} apply deep learning techniques to the image spam problem.
They obtain an accuracy about about~91\%, which is comparable to other research
in the field. 

In addition, to research that deals exclusively with image spam,
there are many articles on spam detection that
cover aspects of the image spam problem. See, for example,~\cite{dada} 
for a recent survey of spam detection research articles.

\section{Learning Techniques}\label{sect:mltechniques}

In this section, we provide background information on the various 
machine learning and neural network techniques considered in this paper. 
Specifically, we discuss support vector machines (SVM), multilayer perceptron (MLP), 
and convolutional neural networks (CNN).

\subsection{SVM}

Support vector machines (SVM) are a class of supervised machine learning algorithms 
that have been extensively studied in the context of email spam in general~\cite{email_svm},
and image spam in particular~\cite{particle_swarm}. In this section, a short overview of SVM is given.


The following four ideas are key to understanding SVM~\cite{app_security}.

\begin{description}

\item[Separating hyperplane] --- In the training phase, an SVM attempts to find a 
hyperplane that acts as the decision boundary between different classes. Of course,
such a hyperplane need no exist, which leads us to consider higher dimensional spaces, 
as discussed below.

\item[Maximize the margin] --- If we can separate the classes using a hyperplane, there will be 
an infinite number of such hyperplanes. In an SVM, we choose the hyperplane that maximizes the 
margin, where margin is defined as the minimum distance between the hyperplane 
and either class of data. An example illustrating such a separating hyperplane---and
the corresponding support vectors---is given in Figure~\ref{fig:svm}.

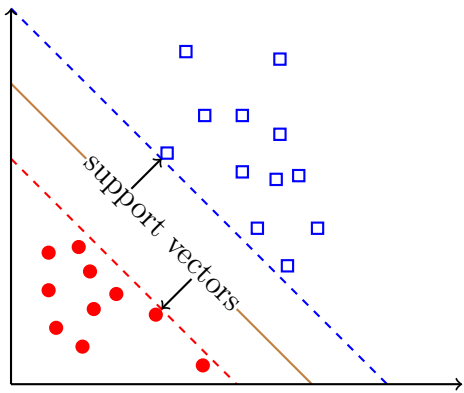
\begin{figure}[!htb]
\centering
    \begin{tikzpicture}[scale=1.0]
    
    \draw[thick,color=blue] (4,2) rectangle (4.15,2.15);
    \draw[thick,color=blue] (3.5,4.25) rectangle (3.65,4.4);
    \draw[thick,color=blue] (3.2,2.0) rectangle (3.35,2.15);
    \draw[thick,color=blue] (3.0,2.75) rectangle (3.15,2.9);
    \draw[thick,color=blue] (3.45,2.65) rectangle (3.6,2.8);
    \draw[thick,color=blue] (3.75,2.7) rectangle (3.9,2.85);
    \draw[thick,color=blue] (3.5,3.25) rectangle (3.65,3.4);
    \draw[thick,color=blue] (3.0,3.5) rectangle (3.15,3.65);
    \draw[thick,color=blue] (2,3) rectangle (2.15,3.15);
    \draw[thick,color=blue] (2.5,3.5) rectangle (2.65,3.65);
    \draw[thick,color=blue] (2.25,4.35) rectangle (2.4,4.5);
    \draw[thick,color=blue] (3.6,1.5) rectangle (3.75,1.65);
    
    \draw[thick,color=red,fill=red] (1.4,1.2) circle (0.08);
    \draw[thick,color=red,fill=red] (0.6,0.75) circle (0.08);
    \draw[thick,color=red,fill=red] (0.95,0.5) circle (0.08);
    \draw[thick,color=red,fill=red] (1.925,0.925) circle (0.08);
    \draw[thick,color=red,fill=red] (0.5,1.25) circle (0.08);
    \draw[thick,color=red,fill=red] (1.1,1.0) circle (0.08);
    \draw[thick,color=red,fill=red] (2.55,0.25) circle (0.08);
    \draw[thick,color=red,fill=red] (1.05,1.5) circle (0.08);
    \draw[thick,color=red,fill=red] (0.9,1.825) circle (0.08);
    \draw[thick,color=red,fill=red] (0.5,1.75) circle (0.08);
    

    
    \draw[thick,color=brown] (0,4) -- (1,3); 
    \draw[thick,color=brown] (3,1) -- (4,0); 
    \draw[thick,dashed,color=blue] (0,5) -- (5,0); 
    \draw[thick,dashed,color=red] (0,3) -- (3,0); 

    \draw[thick,color=black,->] (2.4,1.4) -- (2,1); 
    \draw[thick,color=black,->] (1.6,2.6) -- (2,3); 
    

    \node[rotate=-45] at (2,2){support vectors};
    
     \draw[thick,color=black,->] (0,0) -- (6,0); 
     \draw[thick,color=black,->] (0,0) -- (0,5); 
   
    \end{tikzpicture}
\caption{Support vectors in SVM~\protect\cite{StampML2017}}\label{fig:svm}
\end{figure}

\item[Work in higher dimensions] --- There is no assurance
that the data in the classes will be linearly separable. By transforming the input data to a 
higher dimensional feature space, there is more space to work in, and hence a better chance of 
finding a separating hyperplane. The potential down side to such an approach is that 
computations become more costly. 

\item[Kernel trick] --- The kernel trick enables us to work in a higher dimensional feature space
without paying any significant penalty in terms of computational overhead. By carefully choosing
our kernel transformation, we do not need to explicitly transform our data to a higher dimensional 
space, yet that is precisely what happens behind the scenes.
		
\end{description}

For more information on SVMs, see, for example~\cite{app_security}. As mentioned above,
SVMs have been applied to the image spam problem in~\cite{image_spam,spam_svm_chavda}.

\subsection{MLP}

Neural networks provide a powerful general framework for 
dealing with many challenging learning problems. 
Neural networks can be viewed as modeling neurons in the brain and, as 
with SVMs, neural networks are supervised algorithms
that are suitable for binary or multiclass classification.

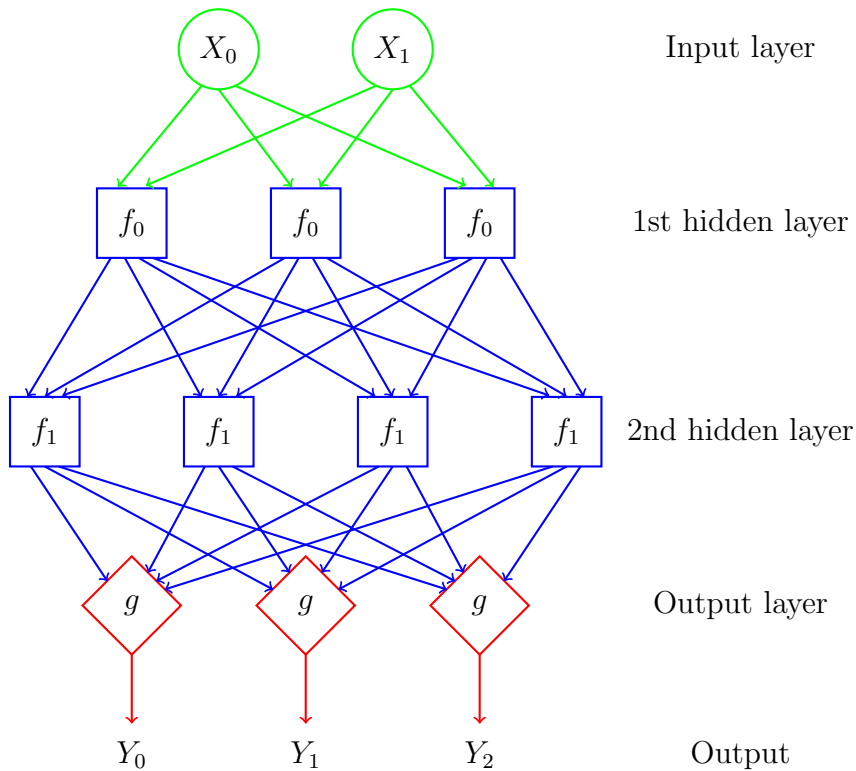
\begin{figure}[!htb]
  \centering
    \begin{tikzpicture}[scale=0.925]
    
    \draw[thick,color=green] (3.0,8.5) circle (0.575);
    \draw[thick,color=green] (5.5,8.5) circle (0.575);

    \draw[thick,color=blue] (1.25,5.5) rectangle (2.25,6.5);
    \draw[thick,color=blue] (3.75,5.5) rectangle (4.75,6.5);
    \draw[thick,color=blue] (6.25,5.5) rectangle (7.25,6.5);
    
    \draw[thick,color=blue] (0.0,2.5) rectangle (1.0,3.5);
    \draw[thick,color=blue] (2.5,2.5) rectangle (3.5,3.5);
    \draw[thick,color=blue] (5.0,2.5) rectangle (6.0,3.5);
    \draw[thick,color=blue] (7.5,2.5) rectangle (8.5,3.5);
    
    \draw[thick,color=red,rotate around={45:(1.75,0.5)}] (1.25,0.0) rectangle (2.25,1.0);
    \draw[thick,color=red,rotate around={45:(4.25,0.5)}] (3.75,0.0) rectangle (4.75,1.0);
    \draw[thick,color=red,rotate around={45:(6.75,0.5)}] (6.25,0.0) rectangle (7.25,1.0);

    \draw[thick,color=green,->] (2.75,7.97) -- (1.55,6.52);
    \draw[thick,color=green,->] (3,7.92) -- (4.05,6.52);
    \draw[thick,color=green,->] (3.25,7.97) -- (6.55,6.54);
    
    \draw[thick,color=green,->] (5.25,7.97) -- (1.95,6.54);
    \draw[thick,color=green,->] (5.5,7.92) -- (4.45,6.52);
    \draw[thick,color=green,->] (5.75,7.97) -- (6.95,6.52);

    \draw[thick,color=blue,->] (0.3,2.5) -- (1.4,0.85);
    \draw[thick,color=blue,->] (0.5,2.5) -- (3.78,0.73);
    \draw[thick,color=blue,->] (0.7,2.5) -- (6.28,0.73);
    
    \draw[thick,color=blue,->] (2.8,2.5) -- (1.98,0.97);
    \draw[thick,color=blue,->] (3.0,2.5) -- (4.02,0.97);
    \draw[thick,color=blue,->] (3.2,2.5) -- (6.4,0.85);
    
    \draw[thick,color=blue,->] (5.3,2.5) -- (2.1,0.85);
    \draw[thick,color=blue,->] (5.5,2.5) -- (4.48,0.97);
    \draw[thick,color=blue,->] (5.7,2.5) -- (6.52,0.97);
    
    \draw[thick,color=blue,->] (7.8,2.5) -- (2.22,0.73);
    \draw[thick,color=blue,->] (8.0,2.5) -- (4.72,0.73);
    \draw[thick,color=blue,->] (8.2,2.5) -- (7.1,0.85);

    \draw[thick,color=blue,->] (1.45,5.5) -- (0.25,3.5);
    \draw[thick,color=blue,->] (1.65,5.5) -- (2.75,3.5);
    \draw[thick,color=blue,->] (1.85,5.5) -- (5.25,3.51);
    \draw[thick,color=blue,->] (2.05,5.5) -- (7.75,3.52);

    \draw[thick,color=blue,->] (3.95,5.5) -- (0.5,3.51);
    \draw[thick,color=blue,->] (4.15,5.5) -- (3.0,3.5);
    \draw[thick,color=blue,->] (4.35,5.5) -- (5.5,3.5);
    \draw[thick,color=blue,->] (4.55,5.5) -- (8.0,3.51);

    \draw[thick,color=blue,->] (6.45,5.5) -- (0.75,3.52);
    \draw[thick,color=blue,->] (6.65,5.5) -- (3.25,3.51);
    \draw[thick,color=blue,->] (6.85,5.5) -- (5.75,3.5);
    \draw[thick,color=blue,->] (7.05,5.5) -- (8.25,3.5);

    \draw[thick,color=red,->] (1.75,-0.2) -- (1.75,-1.2);
    \draw[thick,color=red,->] (4.25,-0.2) -- (4.25,-1.2);
    \draw[thick,color=red,->] (6.75,-0.2) -- (6.75,-1.2);

    \node at (3.0,8.5) {$X_0$};
    \node at (5.5,8.5) {$X_1$};

    \node at (1.75,6.0) {$f_0$};
    \node at (4.25,6.0) {$f_0$};
    \node at (6.75,6.0) {$f_0$};

    \node at (0.5,3.0) {$f_1$};
    \node at (3.0,3.0) {$f_1$};
    \node at (5.5,3.0) {$f_1$};
    \node at (8.0,3.0) {$f_1$};

    \node at (1.75,0.5) {$g$};
    \node at (4.25,0.5) {$g$};
    \node at (6.75,0.5) {$g$};

    \node at (1.75,-1.65) {$Y_0$};
    \node at (4.25,-1.65) {$Y_1$};
    \node at (6.75,-1.65) {$Y_2$};
    
    \node at (10.5,8.5) {Input layer};
    \node at (10.5,6.0) {1st hidden layer};
    \node at (10.5,3.0) {2nd hidden layer};
    \node at (10.5,0.5) {Output layer};
    \node at (10.5,-1.65) {Output};

    \end{tikzpicture}
  \caption{MLP with two hidden layers}\label{fig:FF_nn}
\end{figure}

A perceptron is a simple type of artificial neuron that has an input layer and an output layer.
While conceptually simple, a perceptron is limited to a linear decision boundary, much
like a linear SVM. The equivalent of the kernel trick for perceptrons are multi-layer perceptrons (MLP)
which include one or more hidden layers between the input and the output. An example
of an MLP with two hidden layers is given in Figure~\ref{fig:FF_nn}. The number 
of layers, the number of neurons (i.e., functions) at each layer, and the functions themselves
must be specified as part of an MLP architecture.
Each edge in an MLP graph represents a weight that is learned 
via training. Backpropagation, which is a gradient descent technique,
can be used to efficiently train an MLP~\cite{deep_thoughts}.

Due to the hidden layers, an MLP is not restricted to a linear decision boundary, 
which is very much analogous to an SVM with a nonlinear kernel function. The advantage of
an MLP over a nonlinear SVM is that we do not specify an explicit kernel function---it is as 
if the kernel function of the SVM is learned during training.
However, more training data and computational effort is needed to train 
an MLP---as compared to a nonlinear SVM---since more parameters must be determined.

\subsection{CNN}

Generally, neural networks use fully connected layers,
that is, all neurons at one layer are connected to all neurons in the next layer.
A fully connected layer can deal effectively with correlations 
between any points within the training vectors---regardless of
whether those points are close together, far apart, or somewhere in between. 
In contrast, CNNs 
are designed to deal with local structure, and hence a convolutional layer
cannot be expected to perform well when crucial information is not local. 
The benefit of a CNN is that convolutional layers have far fewer
parameters and hence they can be 
trained much more efficiently than 
fully connected layers.

For images, most of the important structure (edges and gradients, for example) 
is local. Hence, CNNs would seem to be an ideal tool for
image analysis and, in fact, CNNs were developed for precisely this problem.
But, CNNs have performed well in a variety of other problem domains.
In general, any problem for which there exists a data representation
where local structure predominates is a candidate for a CNN.
In addition to images, local structure is key in the fields
of text analysis and speech analysis, among many others.

\begin{figure}[!htb]
  \centering
\begin{tikzpicture}[scale=0.355,every node/.style={scale=0.85}]

\draw[red,ultra thick] (0.0,0.0) rectangle (11.2,11.2);

\draw[green,ultra thick] (0.5,-0.5) rectangle (11.7,10.7);

\draw[blue,ultra thick] (1.0,-1.0) rectangle (12.2,10.2);


\draw[ultra thick, dotted] (0.0,9.1) rectangle (2.1,11.2);
\draw[ultra thick] (0.0,11.2) -- (2.1,11.2);
\draw[ultra thick] (0.0,11.2) -- (0.0,9.1);
\draw[ultra thick] (1.0,8.1) rectangle (3.1,10.2);
\draw[ultra thick] (0.0,9.1) -- (1.0,8.1);
\draw[ultra thick] (2.1,11.2) -- (3.1,10.2);
\draw[ultra thick] (0.0,11.2) -- (1.0,10.2);
\draw[ultra thick, dotted] (3.1,8.1) -- (2.1,9.1);

\draw[black,ultra thick] (13.6,1.4) rectangle (23.4,11.2);
\draw[ultra thick] (13.6,11.2) rectangle (14.3,10.5);
\draw[black,ultra thick] (14.1,0.9) rectangle (23.9,10.7);
\draw[ultra thick] (14.1,10.7) rectangle (14.8,10.0);
\draw[black,ultra thick] (14.6,0.4) rectangle (24.4,10.2);
\draw[ultra thick] (14.6,10.2) rectangle (15.3,9.5);
\draw[black,ultra thick] (15.1,-0.1) rectangle (24.9,9.7);
\draw[ultra thick] (15.1,9.7) rectangle (15.8,9.0);
\draw[black,ultra thick] (15.6,-0.6) rectangle (25.4,9.2);
\draw[ultra thick] (15.6,9.2) rectangle (16.3,8.5);

\draw[smooth,thick,->] (2.5,10.7) -- (2.5,11.9) -- (3.0,12.2) -- (12.95,12.2)  -- (13.55,11.9) -- (13.95,11.2);
\node at (7,12.7) {$\mbox{\footnotesize filter}_1$};
\draw[smooth,thick,->] (2.0,10.7) -- (2.0,12.9) -- (2.5,13.2) -- (12.95,13.2)  -- (13.55,12.9) -- (14.45,10.7);
\node at (7,13.7) {$\mbox{\footnotesize filter}_2$};
\draw[smooth,thick,->] (1.5,10.7) -- (1.5,13.9) -- (2.0,14.2) -- (12.95,14.2)  -- (13.55,13.9) -- (14.95,10.2);
\node at (7,14.7) {$\mbox{\footnotesize filter}_3$};
\draw[smooth,thick,->] (1.0,10.7) -- (1.0,14.9) -- (1.5,15.2) -- (12.95,15.2)  -- (13.55,14.9) -- (15.45,9.7);
\node at (7,15.7) {$\mbox{\footnotesize filter}_4$};
\draw[smooth,thick,->] (0.5,10.7) -- (0.5,15.9) -- (1.0,16.2) -- (12.95,16.2)  -- (13.55,15.9) -- (15.95,9.2);
\node at (7,16.7) {$\mbox{\footnotesize filter}_5$};

\end{tikzpicture}
  \caption{First convolutional layer for RGB image}\label{fig:conv3}
\end{figure}
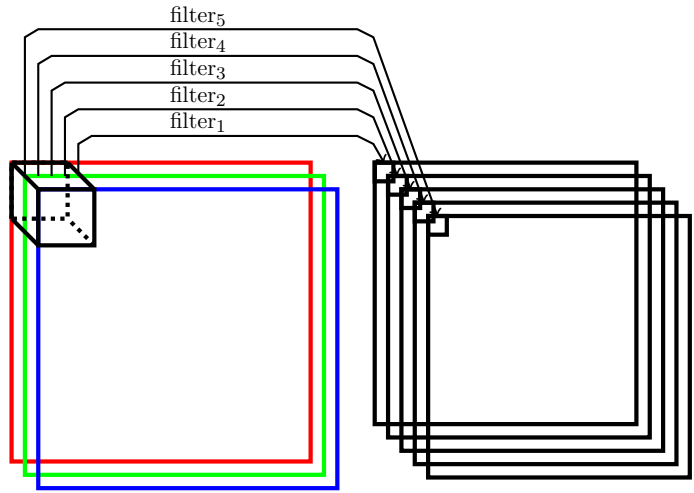

Figure~\ref{fig:conv3} illustrates a simple examples of a convolutional layer.
Note that the convolution is applied to a three-dimensional chunk of image
data (typically, the R, G, and B planes of an RGB image), 
over a sliding window. Also, note that the convolutions are stacked,
which, together with random initialization of the filters, enables the learning of multiple features. 
The example in Figure~\ref{fig:conv3} represents the first convolutional layer, which is applied
directly to the image, whereas subsequent convolutional layers are applied to 
convolutional layers. By applying convolutions to convolutions, higher-level
representations of images can be obtained.

In addition to convolutional layers, CNNs often include pooling layers. Among other things,
pooling layers serve to downsample from a given layer, 
thereby making the subsequent computations more efficient.

\section{Experiments}\label{sect:experiments}

This section presents the results from a variety of experiments. 
But first we discuss the image features used in our experiments,
the criteria we employ to quantify effectiveness, and 
the datasets used in our experiments.

\subsection{Features}

The Canny edge detector is a standard tool for extracting information related to
edges in images~\cite{canny}. The Canny image is a representation of 
an image with its edges highlighted.

\begin{figure}[!htb]
	\begin{center}
  		\includegraphics[scale=0.55]{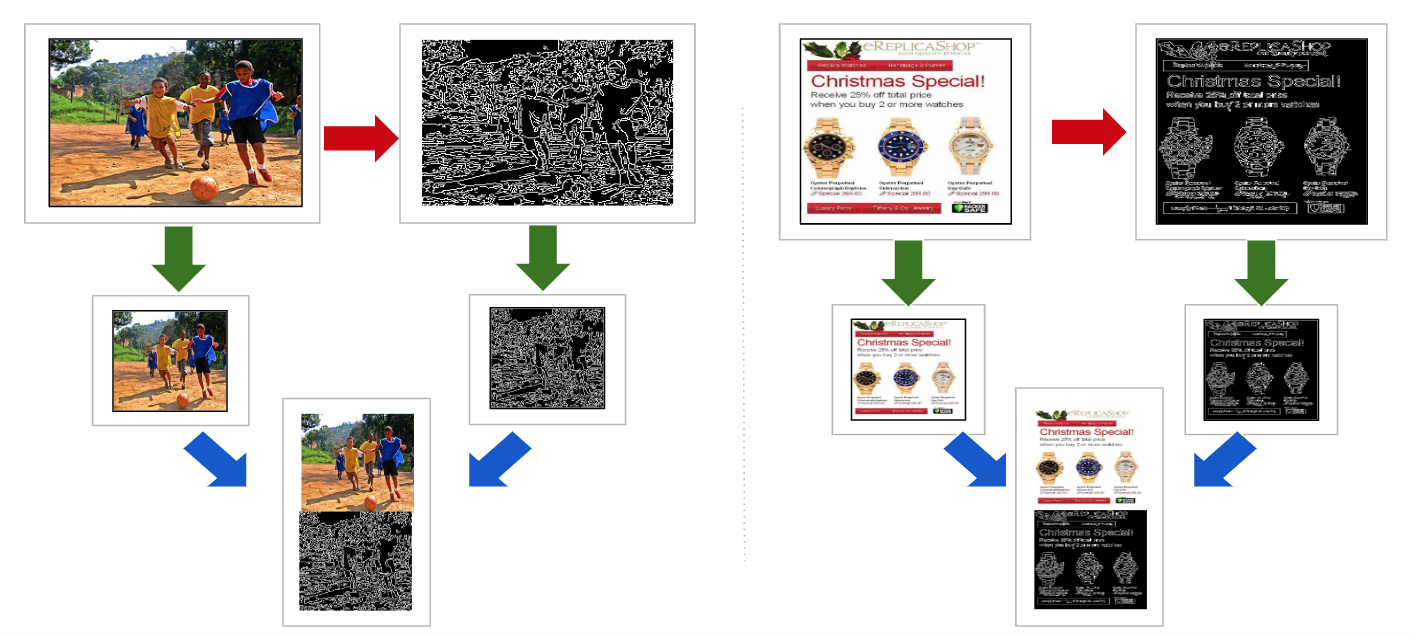}
  		\caption{Raw, Canny, and combination features}\label{feature-gen}
	\end{center}
\end{figure}

For our CNN experiment we consider raw images, as well as Canny images, 
and a combined (or augmented) feature consisting of both the raw and Canny images. 
Examples of raw, Canny, and combined images---for both ham and
spam---are given in Figure~\ref{feature-gen}.

The raw images require no processing other than resizing, while the Canny images
can also be computed efficiently. Note that in comparison to most previous work, 
this feature generation process is extremely efficient. For example,
in~\cite{image_spam} some~21 image features are considered,
while in~\cite{spam_svm_chavda}, a total of~38 features are considered,
and many of these features are at least as computationally intensive 
to extract as the Canny image feature considered here.

\subsection{Evaluation Metrics}

We evaluate our proposed techniques based on accuracy and the area under the ROC curve. 
In the context of image spam detection, true positive (TP) gives the number of correctly 
identified image spam samples, while true negative (TN) is the number of ham images
that are correctly classified as such.
False positive (FP) represents the number of ham images identified as spam,
while false negative (FN) is the number of spam images that are mis-identified as ham. 
Accuracy is given in terms of TP, FP, TN and FN as
$$
\mbox{Accuracy} = \frac{\mbox{TP} + \mbox{TN}}{\mbox{TP} + \mbox{TN} + \mbox{FP} + \mbox{FN}}
$$

To generate a receiver operating characteristic (ROC) curve, we plot the true positive rate versus the 
false positive rate as the threshold passes through the range of scores.
The area under the ROC curve (AUC) is a useful statistic for comparing
performance of classifiers. In the context of image spam detection,
the AUC can be interpreted as the probability that a randomly selected
spam image scores higher than a randomly selected ham image~\cite{auc}.

\subsection{Datasets}

In our experiments, we use three publicly available datasets. One of these
datasets consist of actual image spam collected from the wild, while
the other two are challenge datasets, developed as part of previous research projects.
These challenge datasets are designed to simulate image spam that
is made to look more like ham images. These three datasets are
as follows.

\begin{description}
\item[ISH Dataset] --- 
This dataset was collected by the ``image spam hunter''
group at Northwestern University~\cite{image_spam_hunter}. 
The dataset contains~920 spam images and~810 ham images. All of the images 
are in the {\tt jpg} format. 

\item[Challenge dataset~1] --- 
This challenge dataset was created by the authors of~\cite{image_spam} 
by applying image processing techniques to spam images to make them 
appear more ham-like. The Dredze spam archive public corpus was used as the source of
the spam images~\cite{spam_dredze}. A weighted overlay technique was used
to blend these spam images with the ham images from the ISH dataset. 

\item[Challenge dataset~2] --- 
This challenge dataset was developed as part of the research in~\cite{spam_svm_chavda}
using a different overlay technique than that in~\cite{image_spam}. 
For this dataset, the background of spam images 
was deleted and the resulting image was then overlaid onto a ham image.
This makes the spam text easier to read, as compared to challenge dataset~1, and 
according to the results in~\cite{spam_svm_chavda}, also makes
for a somewhat more challenging detection problem.
\end{description}

\subsection{Environment Setup}

All of our experiments have been performed using an Apple Macbook 
with 8GB of RAM. We use Python to generate the learning models, 
OpenCV for image processing tasks, and the popular 
{\tt scikit-learn} library to implement the machine learning algorithms~\cite{scikit_learn}.
Numpy~\cite{NP} is used for mathematical functions and TensorFlow~\cite{TF} is used
for deep learning training and testing.

Next, we present our experimental results. We we experiment with SVMs, MLPs, and CNNs,
and for each of these learning techniques, we test on each of the three datasets
discussed above. 

\subsection{SVM Experiments}

For our SVM experiments, we need to first construct feature vectors to train the models. 
In the datasets, images are of different sizes. 
Therefore, we first resize all images to~$32\times 32$. 
Next, we use the Canny edge detection method to convert a raw image 
into a Canny image. To build the feature matrix, 
we generate byte data for each pixel in the Canny image. Each pixel 
consists of three bytes, representing the red, green, and blue (RGB) color information 
within the range from~0 to~255. For computational convenience, each numbers 
is normalized to be in the range from~$0$ to~$1$. We also construct a similar feature
vector from the raw byte values (also normalized). Note that each feature vector 
is of length~1024.

For our experiments, we generate separate SVM models for each of the three datasets. 
In each dataset, we perform a random shuffle and use~70\%\ of the image samples for 
training and the remaining~30\%\ for testing. 
In our SVM experiments, we test both linear and RBF kernels and different features.

Table~\ref{tab:svm_init} shows the accuracy of the SVM when trained and tested on the ISH dataset, 
using raw images that have been resized to~$32\times 32$ as compared
to images that are resized to~$16\times 16$. When using the RBF kernel, 
we achieve a best accuracy of~0.9752, which is much better than the
best case for the linear kernel, which is~0.9156 accuracy. 
In both cases, the raw image feature gives better results than the 
Canny image feature. For the RBF kernel, the difference
between the two feature sizes is negligible.




\begin{table}[htb]
\caption{SVM feature size and type comparison (ISH dataset)}\label{tab:svm_init}
\centering
\begin{tabular}{c|cc|cc}\hline\hline
\multirow{2}{*}{Kernel} & \multicolumn{2}{c|}{$32\times 32$ features} 
		& \multicolumn{2}{c}{$16\times 16$ features} \\
            & Raw & Canny & Raw & Canny \\ \hline
RBF    & 0.9748 & 0.9010 & 0.9752 & 0.9048\\ 
Linear & 0.9156 & 0.8492 & 0.8838 & 0.7861\\ \hline\hline
\end{tabular}
\end{table}

\begin{figure}[!htb]
	\centering
	\begin{tabular}{ccc}
  		\includegraphics[scale=0.2]{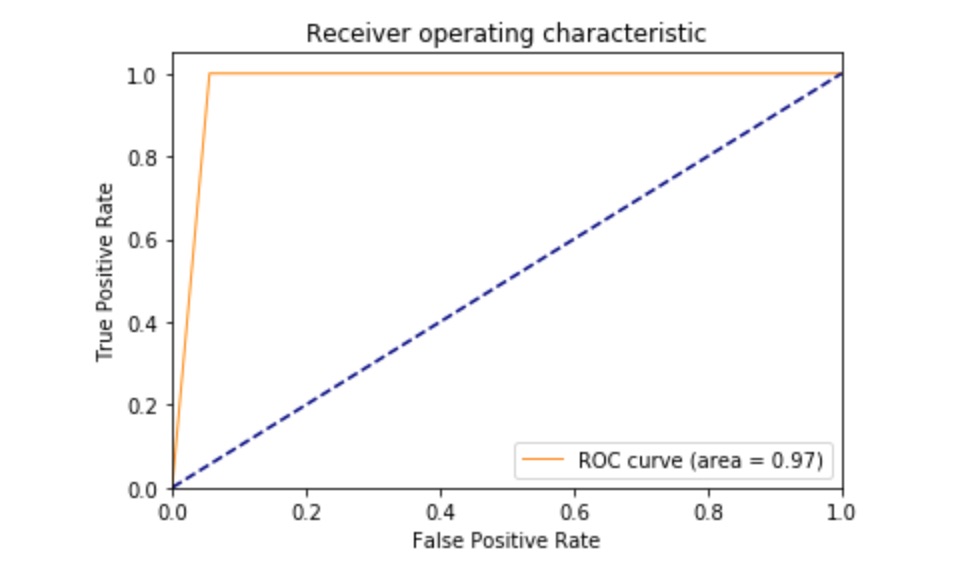}
  		& &
		\includegraphics[scale=0.2]{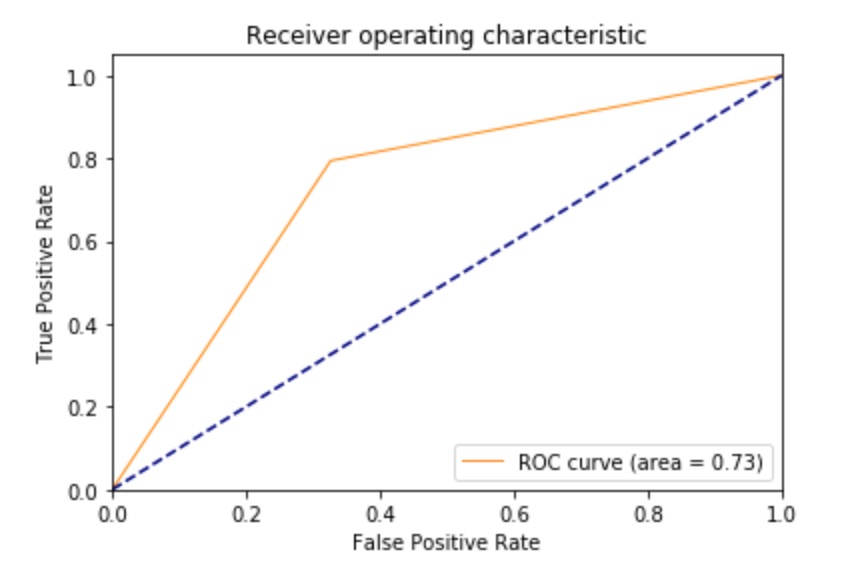} \\
  		(a)~RBF Kernel
		& &
 		(b)~Linear Kernel
	\end{tabular}
 	\caption{ROC curves for ISH dataset}\label{rocsvm}
\end{figure}

Figure~\ref{rocsvm} shows the ROC curves for the SVM binary 
classification results, based on the ISH dataset, 
for both the RBF and linear kernels. 
The corresponding area under the ROC curves (AUC) 
are~0.97 for the RBF kernel and~0.73 for the linear kernel. These results
show that an SVM with RBF kernel can classify ham and spam images 
with high accuracy and a low false positive rate. 

Table~\ref{tab:svm_compare} provides a comparison of the feature types 
(i.e., raw images versus Canny images) and SVM kernel (linear and RBF)
over the three datasets under consideration. We note that the RBF
kernel with raw images performs best in all cases. Hence, we use 
raw images resized to~$16\times 16$ the remaining experiments reported 
in this paper.



\begin{table}[!htb]
\caption{SVM feature and kernel comparison}\label{tab:svm_compare}
\centering
\begin{tabular}{c|cc|cc}\hline\hline
\multirow{2}{*}{Dataset} & \multicolumn{2}{c|}{RBF kernel} & \multicolumn{2}{c}{Linear kernel} \\
   & Raw & Canny  & Raw & Canny\\ \hline
ISH & 0.9752 & 0.9048 & 0.8838 & 0.7861 \\
Challenge~1 & 0.7885 & 0.5553 & 0.4386 & 0.4650 \\ 
Challenge~2 & 0.6715 & 0.6271 & 0.6433 & 0.5965 \\ \hline\hline
\end{tabular}
\end{table}

\begin{figure}[!htb]
	\centering
	\begin{tabular}{ccc}
  		\includegraphics[scale=0.4]{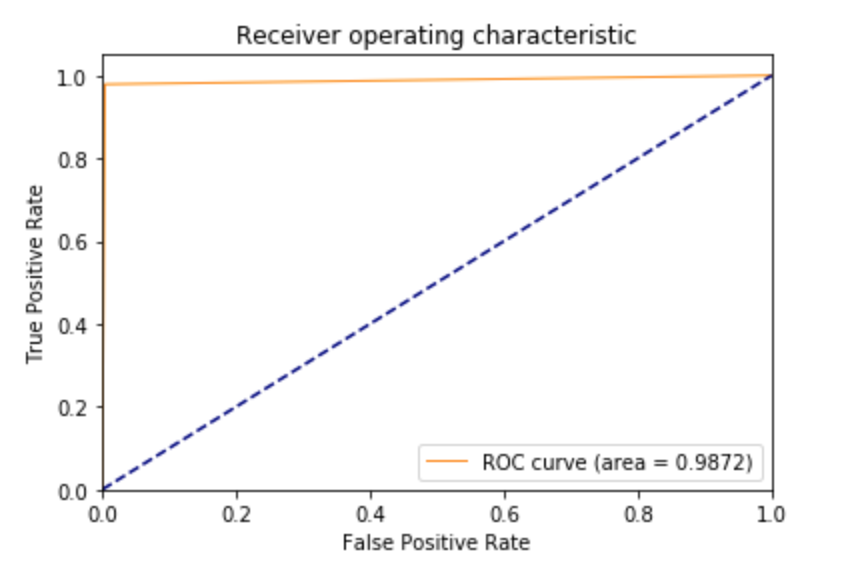}
		&
  		\includegraphics[scale=0.4]{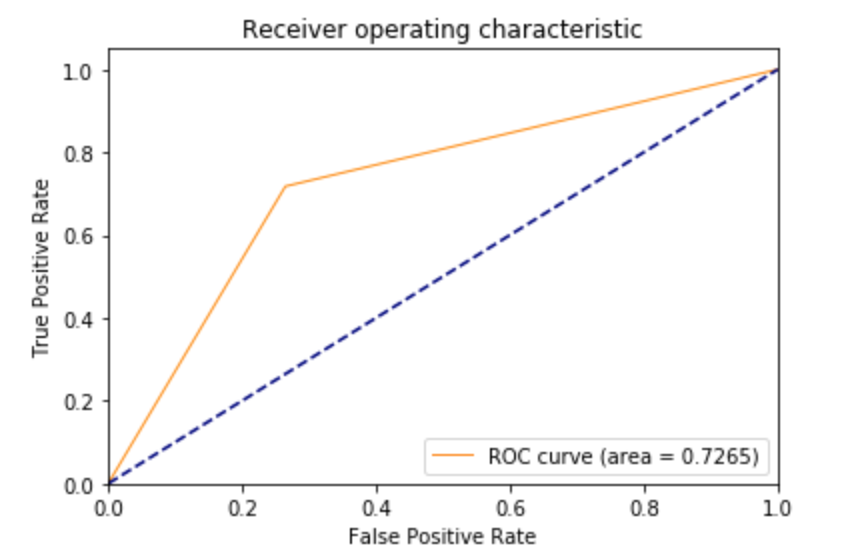}
		\\
 		(a) ISH dataset (RBF)
		&
  		(b) Challenge dataset~1 (RBF)
		\\
		\\
		\multicolumn{2}{c}{\includegraphics[scale=0.4]{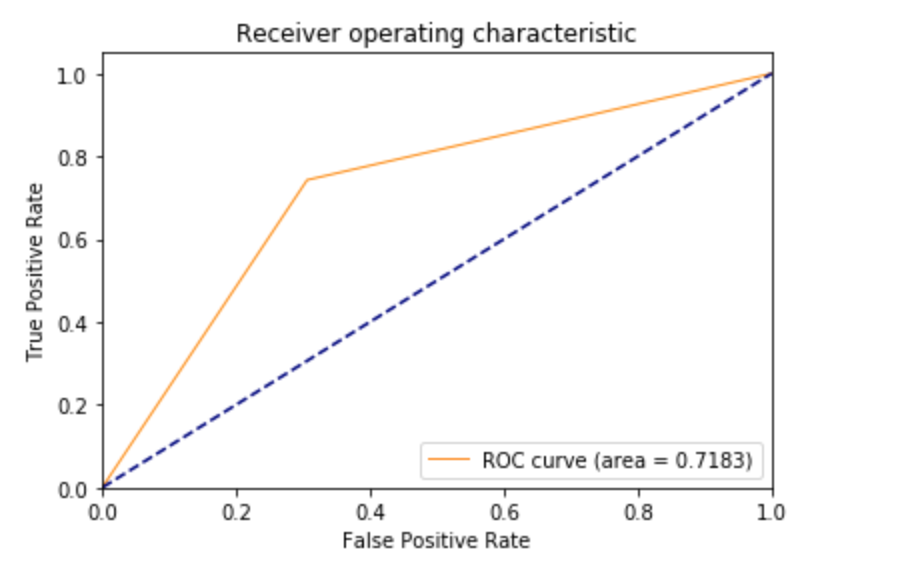}}
		\\
  		\multicolumn{2}{c}{(c) Challenge dataset~2 (Linear)}
	\end{tabular}
  	\caption{ROC curves for combined features}\label{ds_roc}
\end{figure}

Figure~\ref{ds_roc} gives the ROC curves for the best SVM result for each dataset,
based on the combined (raw and Canny) features. As given in Figure~\ref{ds_roc}, 
the AUC values are~0.9872, 0.7265,
and~0.7183 for ISH, challenge~1, and challenge~2 datasets, respectively.
These three cases are all based on the RBF kernel and~$16\times 16$ raw images.

\subsection{MLP Experiments}

For the case of multilayer perceptrons (MLP), we experimented
with several architectures. The results reported here are for MLPs with one input layer, 
two hidden layers, and one output layer. 
Each hidden layer has~128 nodes and 
uses rectifier linear units (ReLU) for the activation functions. To measure the loss, we selected
the binary cross-entropy function. A sigmoid score function is used at the output stage.
This architecture consistently produced the best results.

Our MLPs are trained on~70\%\ of the image samples, and the models are trained for~100 epochs. 
At each epoch, a batch size of~64 is used, and the validation split is taken as~15\%\ of
the image data samples. 

\begin{figure}[!htb]
	\centering
	\begin{tabular}{ccc}
  		\includegraphics[scale=0.375]{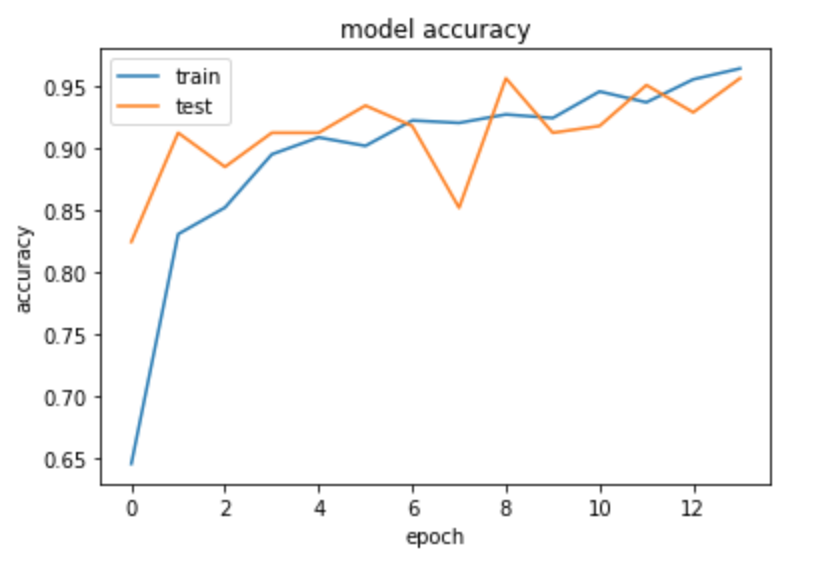}
		& &
  		\includegraphics[scale=0.375]{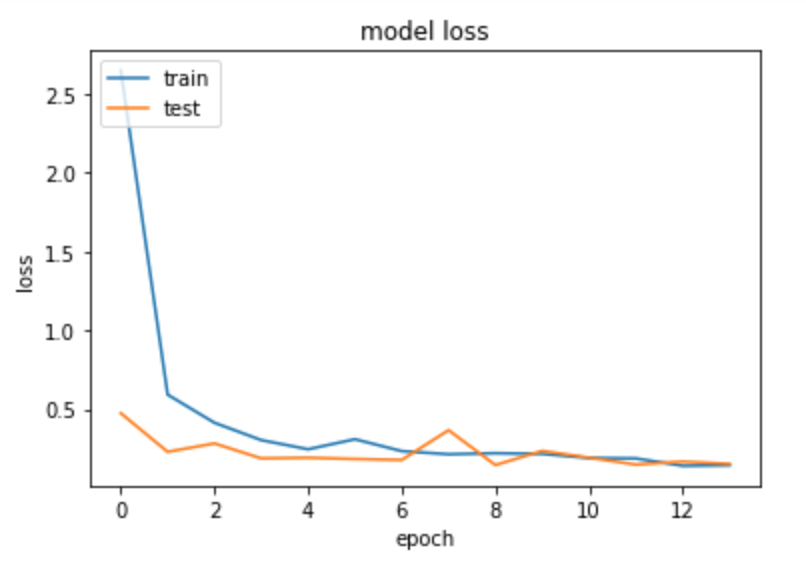} \\
  		(a)~Accuracy for ISH dataset
		& &
  		(b)~Loss for ISH dataset\\
  		\includegraphics[scale=0.375]{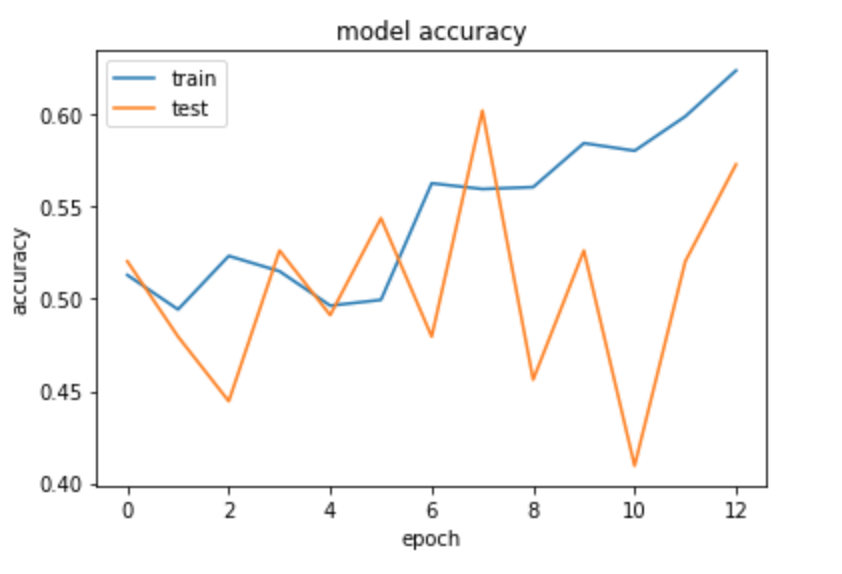}
		& &
  		\includegraphics[scale=0.375]{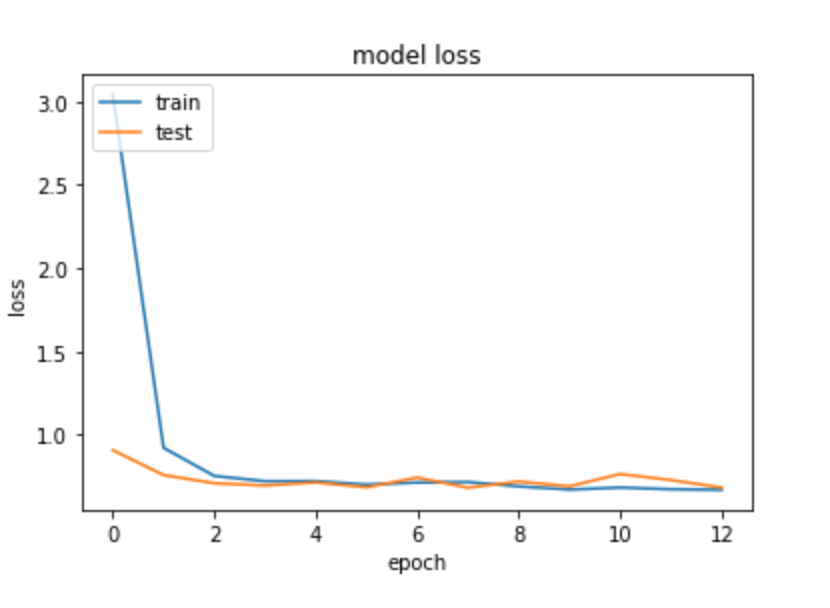} \\
  		(c)~Accuracy challenge dataset~1
		& &
  		(d)~Loss challenge dataset~1\\
  		\includegraphics[scale=0.375]{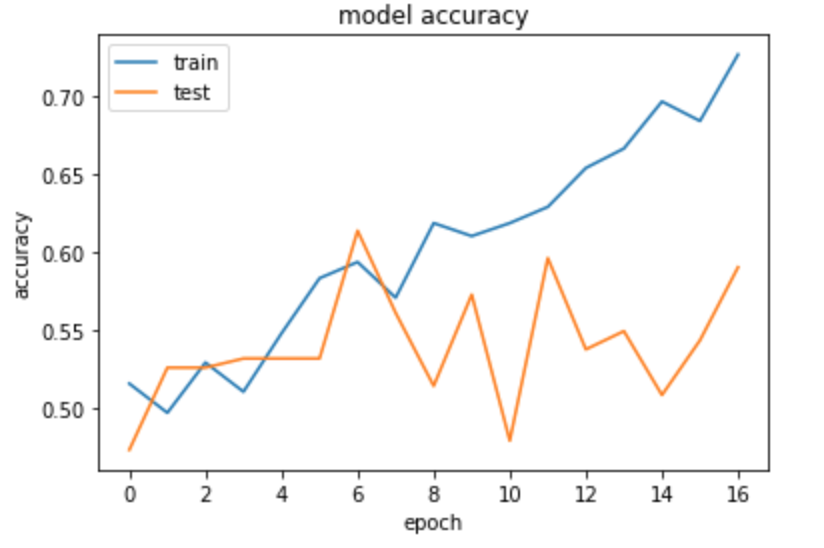}
		& &
  		\includegraphics[scale=0.375]{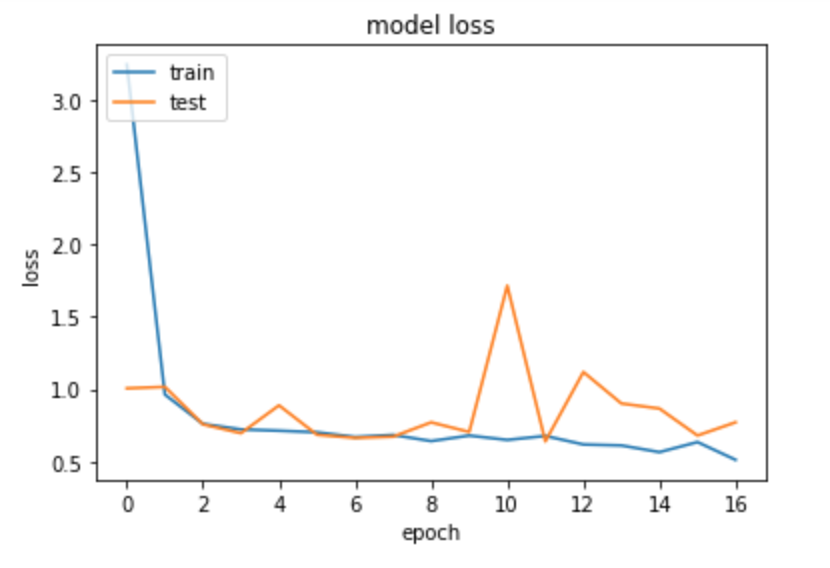} \\
  		(e)~Accuracy challenge dataset~2
		& &
  		(f)~Loss challenge dataset~2
	\end{tabular}
	\caption{MLP accuracy and loss}\label{dsffnnacc_loss}
\end{figure}

Figures~\ref{dsffnnacc_loss}~(a)
and~(b) show the MLP accuracy and loss over~50 epochs, with the 
corresponding loss graph in Figure~\ref{dsffnnacc_loss}~(b).
The analogous MLP accuracy and loss curves for challenge dataset~1 are given
in Figures~\ref{dsffnnacc_loss}~(c) and~(d), while Figures~\ref{dsffnnacc_loss}~(e) and~(f)
give the accuracy and loss results for challenge dataset~2.
These results indicates that in each case, 
the model is converging without overfitting.  

Table~\ref{tab:5} presents the optimal testing accuracies for the MLP experiments 
summarized in Figure~\ref{dsffnnacc_loss}.
In comparison to the SVM results in Table~\ref{tab:svm_compare}, we see that the
MLP fails to outperform the SVM on any of the three datasets.
Also, for challenge dataset~1, the MLP is very poor, performing no better than a coin flip.

\begin{table}[htb]
\caption{MLP results}\label{tab:5}
\centering
\begin{tabular}{c|c}\hline\hline
Dataset & Accuracy \\ \hline
ISH & 0.9557\\
Challenge~1 & 0.5885\\ 
Challenge~2 & 0.6605 \\ \hline\hline
\end{tabular}
\end{table}

\subsection{CNN Experiments}

Convolutional neural networks generally are advantageous---both in terms of 
efficiency and accuracy---for image analysis. As with the SVM and MLP 
experiments discussed above, we apply CNNs 
to each of the three datasets under consideration. 
The results reported in this section are all based on the
combined (raw and Canny image) features.

\begin{figure}[!htb]
\begin{center}
  \includegraphics[scale=0.5]{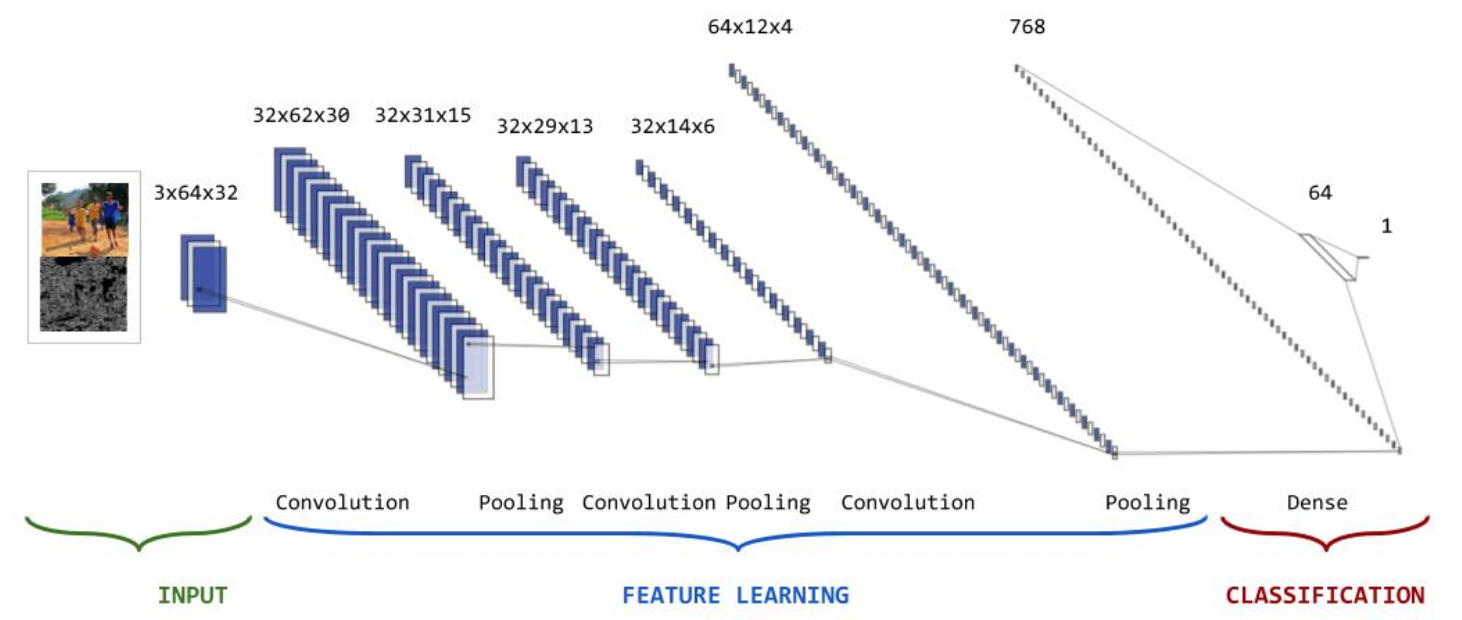}
  \caption{CNN architecture}\label{fig:CNN_arch}
\end{center}
\end{figure}

We experimented with various CNN hyperparameters, but for all of the
experiments reported here, we use the following configuration. 
The first convolution layer uses a~$3\times 3$ kernel and~32 nodes. We have three
convolutional layers, with layers two and three having~32 and~64 nodes, respectively. 
In all convolutional layers, we use the ReLU activation function, and in the final 
fully-connected layer, we use a sigmoid scoring function. We downsample the data 
via a max pooling layer, using a~$2\times 2$ pool size. 
To avoid overfitting, we use a dropout rate of~0.5.
The batch size is set to~64 for each epoch, and we train for~100 epochs
with~70\%\ of the data used for
training and~30\%\ reserved for testing. 
Our CNN architecture is illustrated in Figure~\ref{fig:CNN_arch}.

\begin{figure}[!htb]
	\centering
	\begin{tabular}{ccc}
  		\includegraphics[scale=0.425]{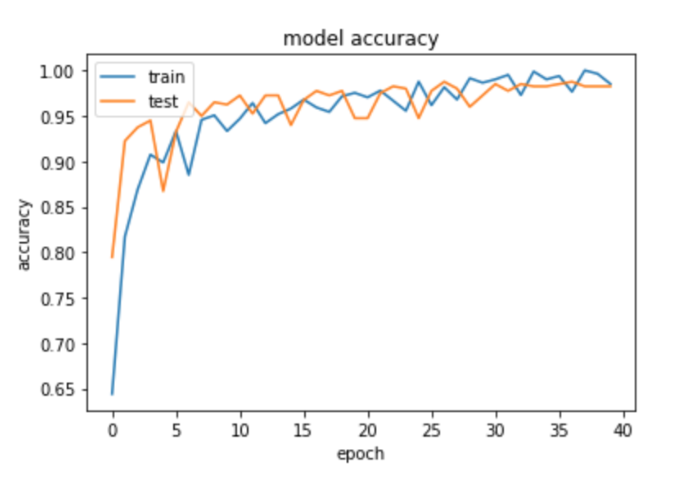}
		& &
  		\includegraphics[scale=0.415]{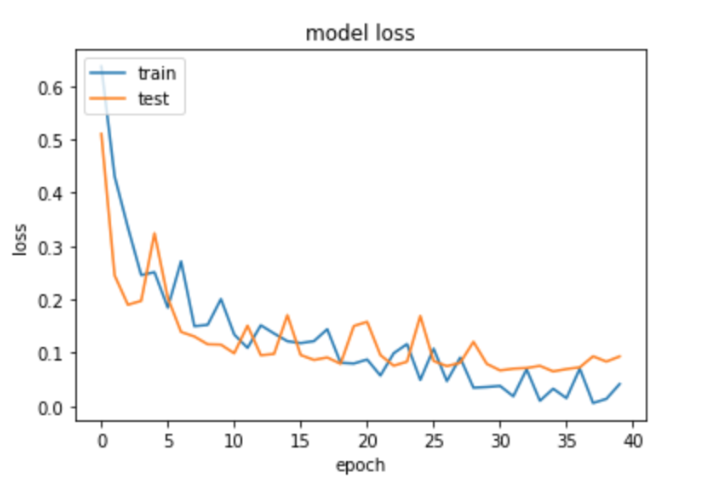} \\
  		(a)~Accuracy for ISH dataset
		& &
  		(b)~Loss for ISH dataset\\
  		\includegraphics[scale=0.35]{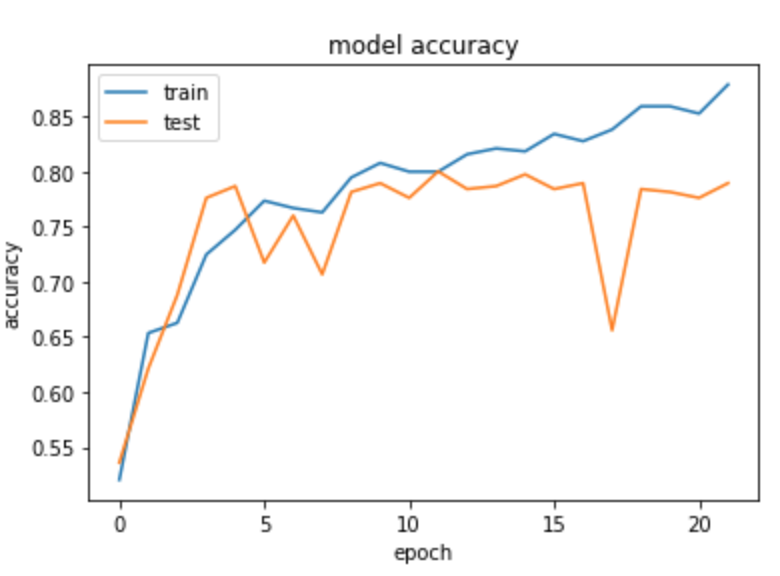}
		& &
  		\includegraphics[scale=0.35]{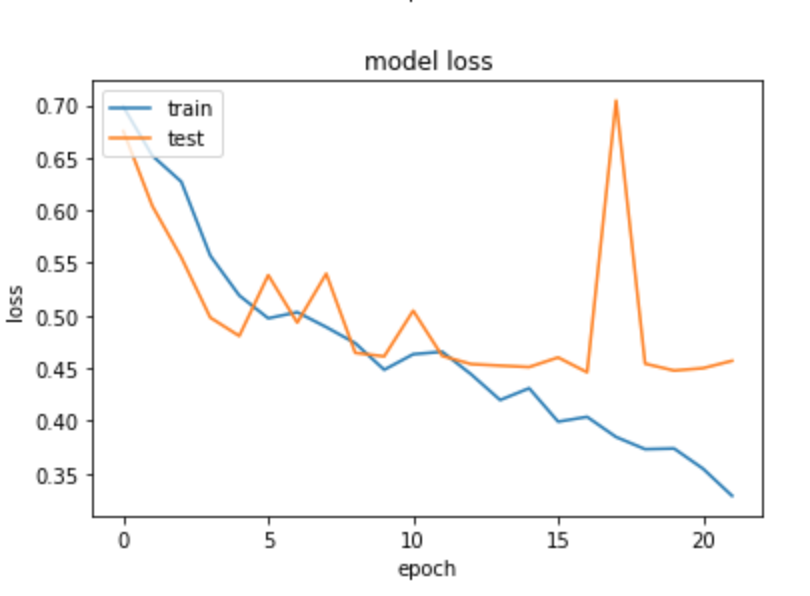} \\
  		(c)~Accuracy challenge dataset~1
		& &
  		(d)~Loss challenge dataset~1\\
  		\includegraphics[scale=0.35]{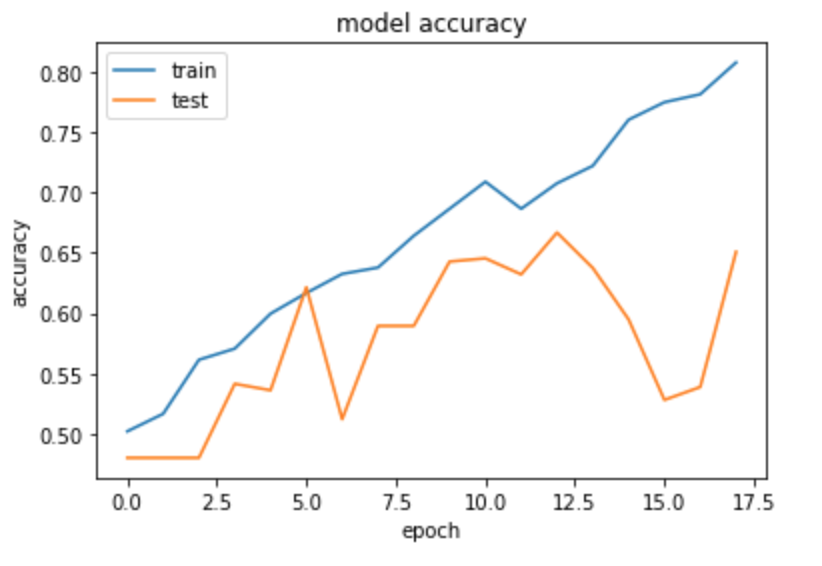}
		& &
  		\includegraphics[scale=0.35]{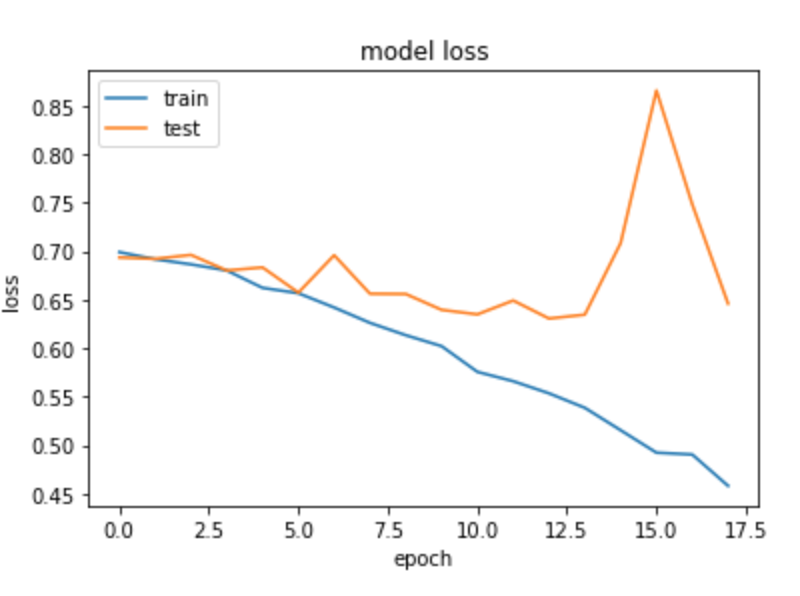} \\
  		(e)~Accuracy challenge dataset~2
		& &
  		(f)~Loss challenge dataset~2
	\end{tabular}
	\caption{CNN accuracy and loss}\label{dscnnacc_loss}
\end{figure}

The accuracy and loss graphs for the ISH dataset
are given in Figures~\ref{dscnnacc_loss}~(a) and~(b), and clearly show that
overfitting does not occur.
The analogous graphs for challenge dataset~1 appears in Figures~\ref{dscnnacc_loss}~(c) and~(d),
while the results for challenge dataset~2 appear in Figures~\ref{dscnnacc_loss}~(e) and~(f).
The spikes that appear in the test graphs for the challenge datasets are indicative of the 
difficulty the models have with the data---even slight improvements on the training set
can result in instability on the validation set. It might be possible to smooth these
spikes somewhat by a heavier use of regularization (e.g., dropouts), but this would not 
otherwise change the results, and would greatly increase the cost of training.

The optimal CNN testing accuracies for the three datasets 
under consideration are given in Table~\ref{tab:6}.
From these results, we see that our CNN outperforms 
both the SVM and MLP on challenge dataset~1, and 
does nearly as well as the SVM on the ISH dataset.

\begin{table}[!htbp]
\caption{CNN results}\label{tab:6}
\centering
\begin{tabular}{c|c}\hline\hline
Dataset & Accuracy \\ \hline
ISH & 0.9902\\ 
Challenge~1 & 0.8313\\ 
Challenge~2 & 0.6769\\ \hline\hline
\end{tabular}
\end{table}

\subsection{Discussion}

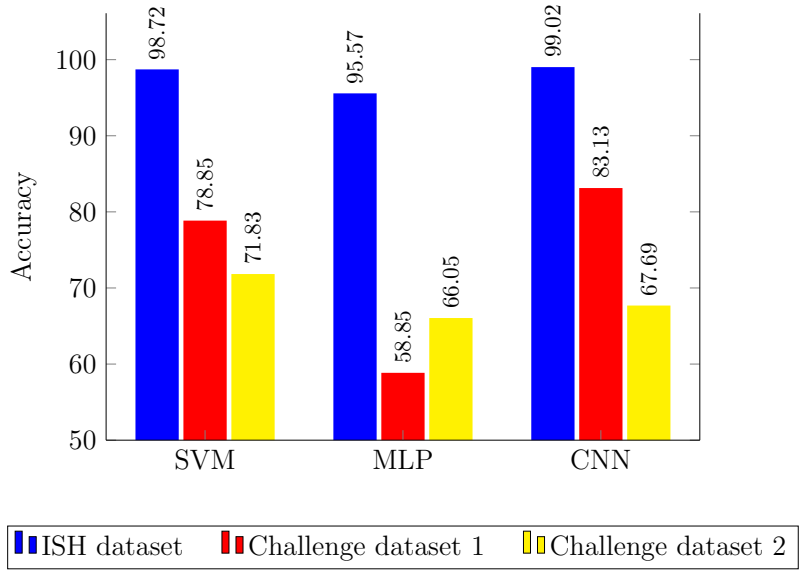
\begin{figure}[!htbp]
\centering
	\begin{tikzpicture}[scale=0.885]
  \begin{axis}[
        ybar, axis on top,
        height=8cm, width=10.5cm,
        bar width=0.65cm,
	tick align=inside,
        enlarge y limits={value=.1,upper},
        ymin=50.0, ymax=101.0,
        axis x line*=bottom,
        y axis line style={opacity=1},
        enlarge x limits=0.25,
        	y tick label style={
    		/pgf/number format/.cd,
   		fixed,
   		fixed zerofill,
    		precision=0},
        nodes near coords,
        every node near coord/.append style={
        		rotate=90, 
		anchor=west, 
		font=\footnotesize,
		/pgf/number format/.cd,
   		fixed,
   		fixed zerofill,
    		precision=2
	},
        legend style={
            at={(0.5,-0.2)},
            anchor=north,
            legend columns=-1,
            /tikz/every even column/.append style={column sep=0.5cm}
        },
        ylabel={Accuracy},
        symbolic x coords={SVM,MLP,CNN},
        xtick=data,
    ]
    \addplot [draw=none, fill=blue] coordinates {
      (SVM,98.72)
      (MLP,95.57) 
      (CNN,99.02)
      };
   \addplot [draw=none,fill=red] coordinates {
      (SVM,78.85)
      (MLP,58.85) 
      (CNN,83.13)
      };
   \addplot [draw=none,fill=yellow] coordinates {
      (SVM,71.83)
      (MLP,66.05) 
      (CNN,67.69)
      };
  \legend{ISH dataset, Challenge dataset~1, Challenge dataset~2}
  \end{axis}
  \end{tikzpicture}
\caption{Comparison of learning techniques}\label{compare_ml}
\end{figure}

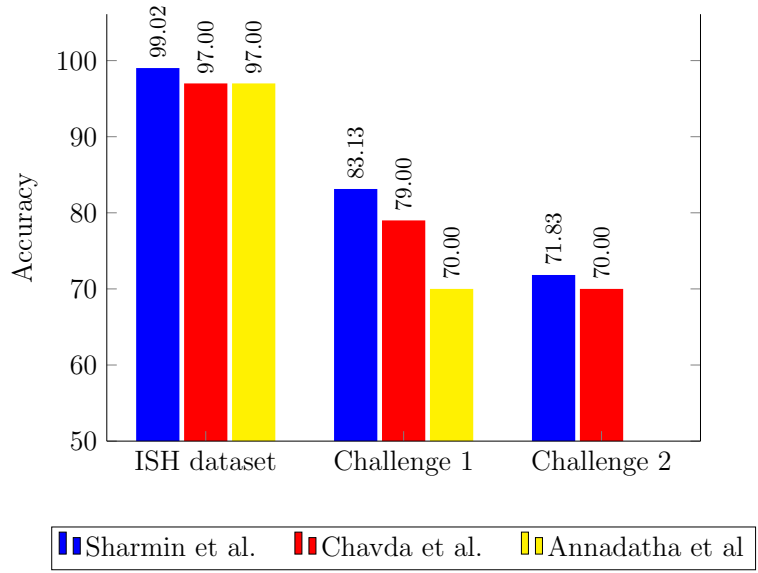
\begin{figure}[!htbp]
\centering
	\begin{tikzpicture}[scale=0.885]
  \begin{axis}[
        ybar, axis on top,
        height=8cm, width=10.5cm,
        bar width=0.65cm,
        tick align=inside,
        enlarge y limits={value=.1,upper},
        ymin=50.0, ymax=101.0,
        axis x line*=bottom,
        y axis line style={opacity=1},
        enlarge x limits=0.25,
        	y tick label style={
    		/pgf/number format/.cd,
   		fixed,
   		fixed zerofill,
    		precision=0},
        nodes near coords,
        every node near coord/.append style={
        		rotate=90, 
		anchor=west, 
		font=\footnotesize,
		/pgf/number format/.cd,
   		fixed,
   		fixed zerofill,
    		precision=2
	},
        legend style={
            at={(0.5,-0.2)},
            anchor=north,
            legend columns=-1,
            /tikz/every even column/.append style={column sep=0.5cm}
        },
        ylabel={Accuracy},
        symbolic x coords={ISH dataset, Challenge~1, Challenge~2},
        xtick=data,
    ]
    \addplot [draw=none, fill=blue] coordinates {
      (ISH dataset,99.02)
      (Challenge~1,83.13)
      (Challenge~2,71.83)
      };
   \addplot [draw=none,fill=red] coordinates {
      (ISH dataset,97.00) 
      (Challenge~1,79.00) 
      (Challenge~2,70.00) 
      };
   \addplot [draw=none,fill=yellow] coordinates {
      (ISH dataset,97.00)
      (Challenge~1,70.00) 
      };
  \legend{Sharmin et al.,Chavda et al.,Annadatha et al}
  \end{axis}
  \end{tikzpicture}
\caption{Comparison to previous work}\label{compare}
\end{figure}

Figures~\ref{compare_ml} and~\ref{compare} provide comparisons of the
various learning techniques presented in this paper, and comparisons
to previous work, respectively.
From Figure~\ref{compare_ml} 
we see that all three machine learning techniques considered in this paper
perform well on the ISH dataset, 
with SVM achieving an accuracy
of~98.72\%\ and a CNN doing even better at~99.02\%, 
while an MLP has a respectable accuracy of~95.57\%. 
For challenge dataset~1, CNN is the clear winner with~83.13\%\ accuracy, 
while none of the techniques can 
perform better than~71.83\%\ accuracy on the more challenging
challenge dataset~2.


In Figure~\ref{compare}, we compare the best results obtained in this research to
that of the related work in~\cite{image_spam} and~\cite{spam_svm_chavda}.
Note that in Figure~\ref{compare}, we refer to the best results in the present paper 
as Sharmin~et~al., while
the work in~\cite{spam_svm_chavda} is denoted as Chavda~et~al., and
the research reported in~\cite{image_spam} is Annadatha~et~al.
From Figure~\ref{compare}, we see that for the ISH dataset, the highest accuracy 
previously achieved was~97\%, while our results top~99\%.
Also, from this same figure, we see that the best result previously obtained
for challenge dataset~1 was~79\%, while we are able to achieve an accuracy 
of more than~83\%\ using a CNN. On challenge dataset~2, we only do marginally
better than the originators of this dataset, Chavda et al.~\cite{spam_svm_chavda}.
Note that dataset~2 was developed after publication of the Annadatha~et~al paper~\cite{image_spam},
which explains the missing bar in Figure~\ref{compare}.


\section{Conclusion}\label{sect:conclusion}

Distinguishing spam images from ham images is an inherently challenging problem. 
In this paper, three machine learning 
techniques were tested---specifically, we  considered support vector machines (SVM)
and two neural network based techniques, namely, multilayer perceptrons (MLP) and 
convolutional neural networks (CNN). We also experimented with features based 
on raw images, Canny images, and a novel combination of the two.
Our experimental results improved on previous related work involving the
same datasets.

Extensive experiments based on three datasets demonstrated the effectiveness of the 
proposed approaches. We found that an SVM model achieved the best accuracy 
on a public image spam dataset, while a CNN technique performed best on an image spam-like
challenge dataset. Our results for the public image spam dataset marginally exceed
those obtained in previous work, while our results for the previously mentioned challenge
dataset are substantially better than any previous work. 

Furthermore, since our feature set consists only of (resized) raw images and Canny images,
our feature extraction and scoring processes are extremely efficient in comparison
to previous work. 
In most previous work, a large number of image features are required
to achieve results comparable to those obtained here. For example, the
authors of~\cite{image_spam} consider~21 image features, while the work in~\cite{spam_svm_chavda}
is based on a set of~38 features. Yet, Figure~\ref{compare} shows that the
work presented here outperforms the results given in both of these earlier papers.

Future work will include experimenting with additional combinations of features and
hyperparameters. In particular, additional CNN experiments involving Canny images 
would seem to be a promising path to pursue. Deep learning techniques 
such as recurrent neural networks (RNN) and long short term memory (LSTM)
would also be interesting to study in the context of image spam detection.

\bibliographystyle{apacite}

\bibliography{references.bib,Stamp-Mark.bib}

\end{document}